\documentclass[10pt,twocolumn,letterpaper]{article}

\usepackage{iccv}              

\usepackage{microtype}
\usepackage{graphicx}
\usepackage{booktabs} 

\usepackage{amsmath}
\usepackage{amssymb}
\usepackage{mathtools}
\usepackage{amsthm}
\usepackage{pifont}
\usepackage{marvosym}
\usepackage{subcaption}
\usepackage{multirow}

\usepackage{tikz}
\usetikzlibrary{arrows.meta, positioning, shapes.geometric}

\usepackage{pifont}
\usepackage{amssymb}
\usepackage{makecell}
\usepackage{multicol}

\def\eg{\emph{e.g.,}}

\newcommand{\Ours}{\texttt{MovieAgent}\xspace}

\definecolor{iccvblue}{rgb}{0.21,0.49,0.74}
\usepackage[pagebackref,breaklinks,colorlinks,allcolors=iccvblue]{hyperref}

\def\paperID{5058} 
\def\confName{ICCV}
\def\confYear{2025}

\title{Automated Movie Generation via Multi-Agent CoT Planning}

\author{
{\large
Weijia Wu },
{\large
Zeyu Zhu },
{\large
Mike Zheng Shou$^{({\textrm{\Letter}})}$}\\
\\
{\large
Show Lab, National University of Singapore$ \qquad $}
}

\begin{document}
\maketitle
\let\thefootnote\relax\footnotetext{$^{\textrm{\Letter}}$ Corresponding author.}
\begin{abstract}
Existing long-form video generation frameworks lack automated planning, requiring manual input for storylines, scenes, cinematography, and character interactions, resulting in high costs and inefficiencies.
To address these challenges, we present \Ours, an automated movie generation via multi-agent Chain of Thought (CoT) planning. 
\Ours offers two key advantages:
1) We firstly explore and define the paradigm of automated movie/long-video generation.
Given a script and character bank, our \Ours can generates multi-scene, multi-shot long-form videos with a coherent narrative, while ensuring character consistency, synchronized subtitles, and stable audio throughout the film.
2) \Ours introduces a hierarchical CoT-based reasoning process to automatically structure scenes, camera settings, and cinematography, significantly reducing human effort. 
By employing multiple LLM agents to simulate the roles of a director, screenwriter, storyboard artist, and location manager, \Ours streamlines the production pipeline.
Experiments demonstrate that \Ours achieves new state-of-the-art results in script faithfulness, character consistency, and narrative coherence.
Our hierarchical framework takes a step forward and provides new insights into fully automated movie generation.
The code and project website are available at: \href{https://github.com/showlab/MovieAgent}{\color{blue}{\tt Code}} and \href{https://weijiawu.github.io/MovieAgent/}{\color{blue}{\tt Website}}.


\end{abstract}    
\section{Introduction}
\label{sec:intro}

\begin{quote}
    ``\textit{Every great movie should seem new every time you see it.}''  
    \begin{flushright}
        — Roger Ebert
    \end{flushright}
\end{quote}

Automated movie generation creates long-form videos with consistent characters, synchronized subtitles, and audio, given a script synopsis and character bank.
It involves automating narrative planning, scene structuring, and shot composition, replicating the hierarchical reasoning of real-world filmmaking.
Most existing video generation research~\cite{svd,zhou2022magicvideo,imagenvideo,wu2023tune} still focuses on short video generation without structured narratives, such as diffusion-based models like Stable Video Diffusion\cite{svd}, Video LDM\cite{blattmann2023align}, and I2VGen-XL\cite{zhang2023i2vgen}. 
More recently, spatiotemporal transformer models, including Sora\cite{videoworldsimulators2024} and HunyuanVideo~\cite{kong2024hunyuanvideo}, have demonstrated superior performance in generating high-quality short videos (within 10 seconds) with realistic visuals and smoother motion dynamics.
Compared to short-video generation, the development of long-form video generation~\cite{wu2024moviebench,nuwaxl,hu2024storyagent,polyak2024movie} has been relatively slow and still faces many challenges, such as maintaining narrative coherence, character consistency, structured scene transitions, and synchronized audio.
DreamFactory\cite{xie2024dreamfactory} uses multi-agent systems and video generation models to synthesize keyframes, later expanded into long-form videos. 
Similarly, StoryAgent\cite{hu2024storyagent} employs multiple agents for customized storytelling video generation.
However, these approaches are limited to basic long-video synthesis, lacking high-level planning and logically structured multi-scene narratives.
They also fail to handle multi-object interactions, customization, and audio consistency, making them unsuitable for real-world applications.
Thus, automated movie-level long-form generation remains an open challenge in the field.

\begin{figure*}[t]
	\includegraphics[width=0.99\linewidth]{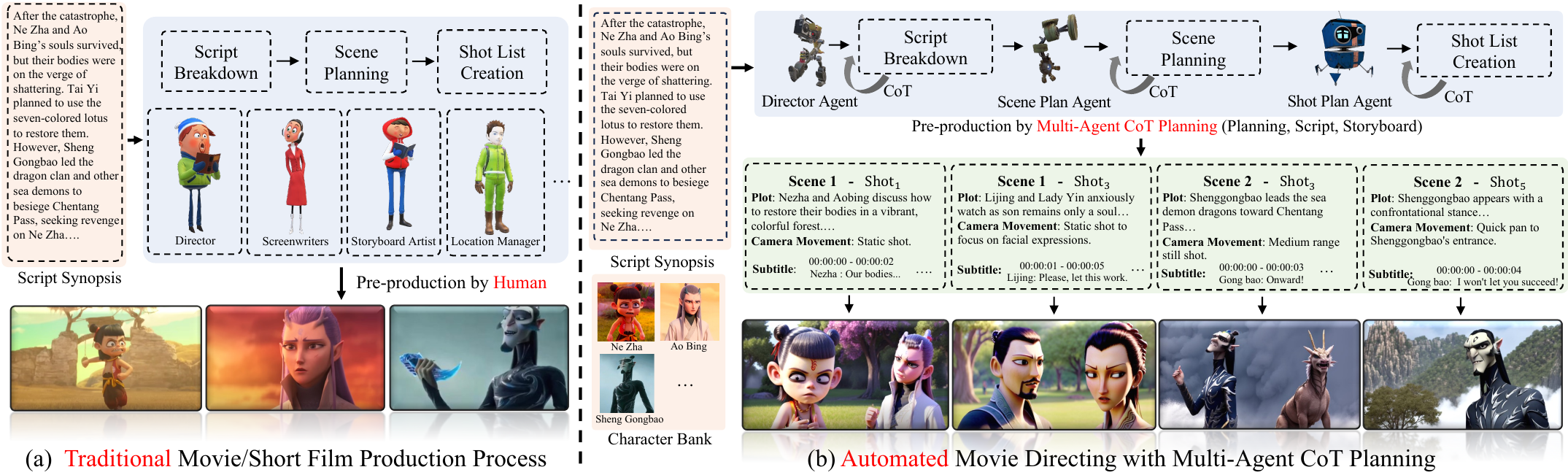}
	\vspace{-0.2cm}
	\caption{\textbf{Comparison of Traditional and Automated Movie Production.} 
     Traditional filmmaking requires manual planning, while our \Ours automates script breakdown, scene planning, and shot design, enhancing efficiency and narrative coherence.
    }
    \vspace{-0.1cm}
\label{motivation}
\end{figure*}

Let us delve deeper into understanding what is essential and indispensable in the real-world movie production process, as shown in Figure~\ref{motivation} (a).
In reality, real-world movie production is a \textbf{hierarchical} and \textbf{collaborative} process, involving multiple specialized roles: directors, screenwriters, storyboard artists, and cinematographers, who work together to maintain narrative coherence, character consistency, and structured scene transitions.
Therefore, unlike short-video generation, movie level video generation is a complex process, including high-level cinematic themes and low-level cinematographic parameters, making it difficult to solve with a single model like an LLM or video generation framework.
%


%
%
%

Inspired by the real-world movie production process, we introduce multi-agent systems to simulate the roles of different filmmaking professionals and implement a hierarchical reasoning framework.
As shown in Figure~\ref{motivation} (b), we propose \Ours, a Multi-Agent CoT Planning framework that automatically structures and generates multi-scene, multi-shot videos with logical storylines, synchronized subtitles, and consistent character appearances.
The key advantages of \Ours include:
1) \textbf{CoT-based Hierarchical Reasoning}. 
Unlike direct inference, which lacks structured and in-depth planning, CoT-based reasoning enables step-by-step, interpretable decision-making while recording the rationale behind each decision for use in subsequent steps.
2) \textbf{Near-zero Cost.} Compared to real-world movie production, which requires millions of dollars and over five years to complete, AI-driven movie generation~(\Ours) is virtually cost-free, as shown in Table~\ref{tab:Costs}.
3) \textbf{Multi-Agent} for Automated Filmmaking. \Ours incorporates multiple specialized AI agents that simulate the roles of a director, screenwriter, and storyboard Artist.
%
Therefore, the multi-agent framework can automatically decomposes a movie synopsis into structured acts, scenes, and shots, ensuring coherent plot development and seamless transitions, as shown in Figure~\ref{motivation} (b).
These agents collaboratively enables precise control over both high-level cinematic themes and low-level cinematographic parameters simultaneously.
%

\begin{table}
    \centering
    \small 
    \setlength{\tabcolsep}{1mm}
    \caption{\textbf{Comparison of Movie/Long-Video Production Costs.} 
    `m', `k', `yrs', and `mins' denotes `million', `thousand', `years', and `minutes', respectively.
    Platform Videos refer to video content for platforms such as YouTube, TikTok.
    Compared to the high-cost, labor-intensive traditional long-video production, \Ours automatically generate video in under $10$ minutes at almost no cost.
    }
\begin{tabular}{c|cc|c|c}
     \multirow{2}{*}{} & \multicolumn{2}{c|}{Traditional Movies} & \multirow{2}{*}{Platform Videos} &
    \multirow{2}{*}{MovieAgent}\\
    \cline{2-3} 
     & Frozen II  & Inside Out 2 &  &  
    \\
    \hline

    Cost & 150m &   200m  & 1 to 100k   & free      \\
    Time & 6 to 7 yrs &  5 to 6 yrs  &  1 to 100 days  & 2 to 10 mins     
\end{tabular}
    \label{tab:Costs}
\end{table}

To summarize, the contributions of this paper are: 
\begin{itemize}[leftmargin=*]
    \item We firstly explore and define the paradigm of automated movie/long-video generation.
    Given a script and character bank, our \Ours can generates multi-scene, multi-shot long-form videos with a coherent narrative, ensuring character consistency, synchronized subtitles.

    \item \Ours employs a hierarchical CoT-based multi-agent reasoning framework to automate scene structuring, camera settings, and cinematography, reducing human effort. 
    With internal CoT reasoning, \Ours effectively decouples and designs cinematic elements, including narrative structure, shot/scene composition, emotional tone, and subtitles.
    
    \item Experiments demonstrate that \Ours achieves state-of-the-art performance in automated storytelling and movie generation.
    %
    %
    Specifically, it excels in character consistency and narrative coherence, providing new insights into fully automated movie generation.

\end{itemize}
\section{Related Works}

\subsection{Video Generation}
Recent advancements in video generation have significantly improved quality and consistency, with approaches spanning diffusion models~\cite{videodiffusionmodel,chen2023seine,zhang2023i2vgen,wu2024draganything,zhao2024motiondirector}, and transformer frameworks~\cite{yan2021videogpt,yang2024cogvideox,kong2024hunyuanvideo}.
Diffusion models have demonstrated remarkable success in image and video synthesis by gradually refining noise into realistic samples.
VDM~\cite{videodiffusionmodel} pioneered the use of diffusion for video generation, introducing a spatiotemporal architecture to model frame dependencies. 
SVD~\cite{svd} further advanced this by leveraging pre-trained text-to-image models for video generation, significantly improving quality.
Lavie~\cite{wang2024lavie} introduces a high-quality video generation framework using cascaded latent diffusion models.
More recently,
SORA~\cite{videoworldsimulators2024} and Hunyuanvideo~\cite{kong2024hunyuanvideo} showcased a highly coherent video generation system using advanced latent diffusion. 
Transformers have emerged as powerful architectures for modeling long-range dependencies. 
VideoPoet~\cite{kondratyuk2023videopoet} introduced a VQ-VAE-based approach, tokenizing video frames and modeling them with an autoregressive transformer. 
CogVideo~\cite{yang2024cogvideox} extended this paradigm with pre-trained text-to-video capabilities, leveraging hierarchical attention mechanisms to improve generation efficiency. 
VideoPoet~\cite{kondratyuk2023videopoet} used a multimodal transformer to improve video-text understanding, enabling more controllable and expressive video synthesis.
Despite advancements, existing frameworks still rely on manual input for narrative planning, cinematography, and scene composition.
Our \Ours addresses these limitations by introducing a multi-agent, where agents simulate key filmmaking roles, enabling fully automated movie generation.

\subsection{Story Visualization}
Story visualization, which generates coherent visual sequences from text, is crucial for automated movie generation.
Early GAN-based methods, such as StoryGAN~\cite{li2019storygan}, focused on maintaining narrative consistency in image sequences. 
With the rise of diffusion models, approaches like StoryDiffusion~\cite{zhou2024storydiffusion}, Magic-Me~\cite{ma2024magic} and DreamVideo~\cite{wei2024dreamvideo} improved temporal coherence and motion dynamics in story-driven videos.
Adapter-based techniques, including IP-Adapter~\cite{ye2023ip}, ROICtrl~\cite{gu2024roictrl} and In-context LoRA~\cite{huang2024context}, enabled efficient fine-tuning for personalized and character-consistent generation.
Meanwhile, structured story-to-video frameworks like AutoStory~\cite{wang2024autostory} and Make-a-story~\cite{rahman2023make} enhanced scene composition and transition planning.
However, existing methods still lack automated high-level planning, often requiring manual intervention for cinematography, scene structuring.
We introduces a multi-agent CoT-driven framework, enabling fully automated and coherent long-form movie generation.

\subsection{LLM for Video Generation}
Recent advancements in LLM-driven video generation~\cite{zhu2023moviefactory,wu2024moviebench} have improved narrative structuring and interactive storytelling.
VideoDirectorGPT~\cite{lin2023videodirectorgpt} and VideoStudio~\cite{long2024videostudio} explored LLM-powered frameworks for scene composition, while Mora~\cite{yuan2024mora} enhanced video conceptualization for long-form coherence. 
For storyboarding and cinematic planning, DreamFactory~\cite{xie2024dreamfactory} and StoryAgent~\cite{hu2024storyagent} introduced LLM-based adaptive shot planning, reducing manual effort in camera control and character interactions. 
VideoGen-of-Thought~\cite{zheng2024videogen} leveraged CoT reasoning to improve multi-shot video consistency. 
Although these methods enhance narrative structuring and storytelling with LLMs, they still require manual intervention or lack character and audio customization.
In this paper, we firstly propose automated movie/long-video generation with a hierarchical CoT reasoning framework, which, given a script, character photos, and audio samples, automates planning, scene structuring, and cinematography for a more coherent and customizable filmmaking process.
\section{Method}

\subsection{Task Formulation}

\label{TaskFormulation}
Given a script synopsis $S$ and a character bank $C$, the goal of automated movie generation is to generate a long-form video $\widehat{\mathcal{V}}$ consisting of multiple scenes and shots while ensuring narrative coherence, character consistency, and audiovisual synchronization.
Formally, the objective is to find an optimal mapping function:
\begin{equation}
\mathcal{F}: (S, C) \rightarrow \widehat{\mathcal{V}}
\end{equation}
where character bank $C = \{[\texttt{char}_k, I_k, A_k]\}_{k=1}^{L}$, $L$ denotes the number of characters, and
$\texttt{char}_k$ is the $k$-th character  names in the character list. 
$I_k$ and $A_k$ denotes the portrait images and audio samples of the character.
The function $\mathcal{F}(\cdot)$ refers to the automated movie generation function that systematically plans sub-scripts, scenes, and shots, along with various shot parameters, camera movements, and cinematographic settings.
Ultimately, it generates a sequence of shots that collectively form the final movie output $\widehat{\mathcal{V}} = \{ \widehat{V}_j^i \mid i = 1, 2, \dots, N, \; j = 1, 2, \dots, M \}$, where $\widehat{V}_j^i$ denotes the $j$-th shot video in the $i$-th scene.



\begin{figure*}[t]
	\includegraphics[width=0.99\linewidth]{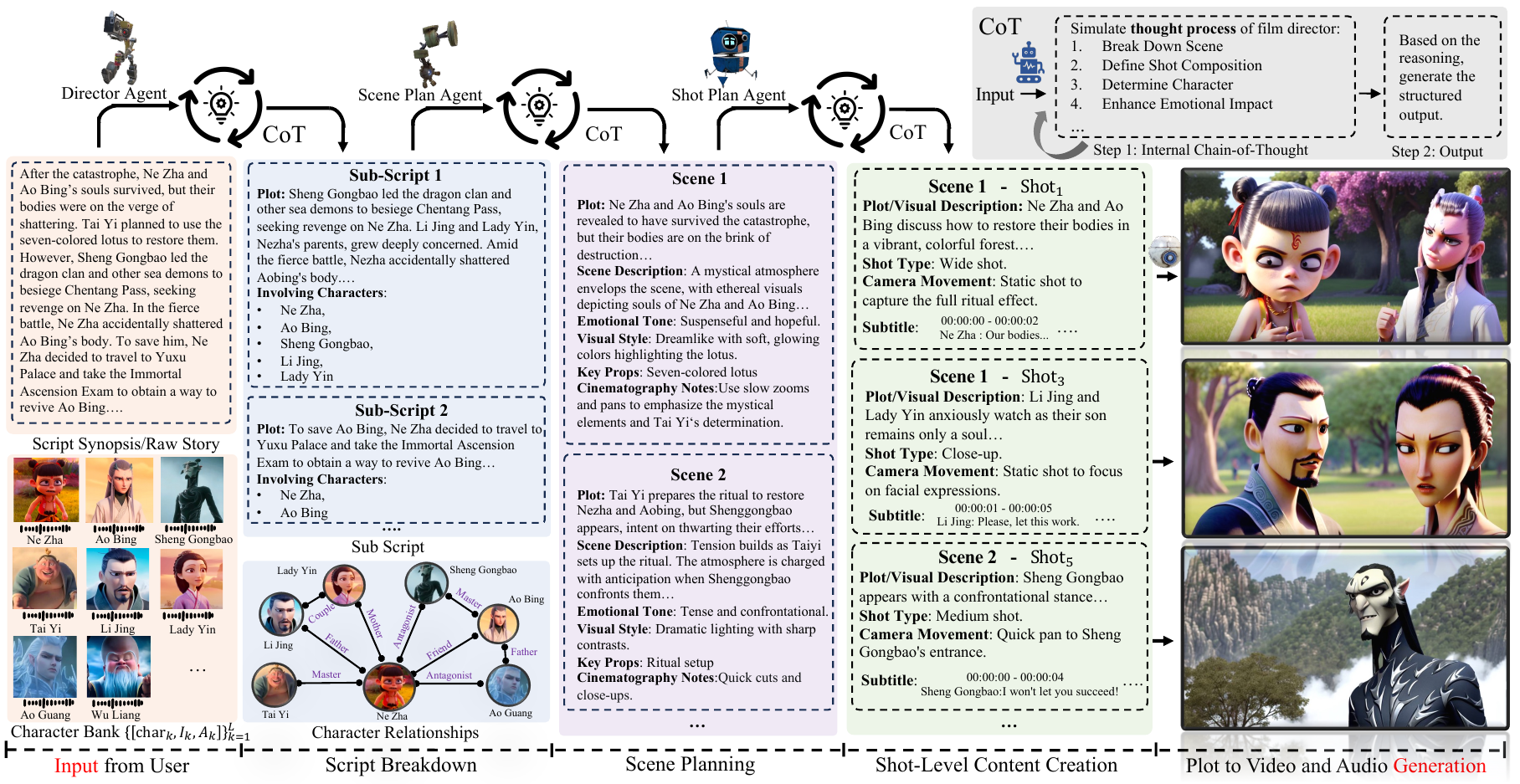}
	\vspace{-0.2cm}
	\caption{\textbf{The Overall Pipeline for \Ours.} 
    The proposed framework employs a hierarchical CoT reasoning process with director, scene plan, and shot plan agents to automate long-form movie generation.
    }
    \vspace{-0.1cm}
\label{Pipeline}
\end{figure*}

\subsection{Automated Movie Generation}



\Ours leverages a multi-agent Chain of Thought reasoning process~(
\S\ref{cottt}) to achieve structured and automated movie generation, as shown in Figure~\ref{Pipeline}.
The system decomposes the filmmaking process into a \textbf{hierarchical workflow}, simulating key roles in traditional movie production. 
Specifically, we introduce three specialized agents: Director Agent~(
\S\ref{Director}), Scene Plan Agent~(
\S\ref{Scene}), and Shot Plan Agent~(
\S\ref{Shot}), which collaboratively structure narratives, plan scenes, and generate detailed cinematographic elements.
Then, customized shot and audio generation~(
\S\ref{videogen}) is utilized to produce the final audio and video.

\subsubsection{Director Agent}
\label{Director}
The Director Agent is responsible for high-level narrative structuring.
Given a script synopsis $S$ and a character bank $C$, it systematically decomposes the storyline into sub-scripts $\mathcal{S} = \{S_1, S_2, ..., S_K\}$, where each $S_p$ represents $p$-th key narrative unit that contributes to the overall plot development.
The segmentation function can be formulated as:
\begin{equation}
    \mathcal{S} = \mathcal{F}_{\text{Director}}(S, C, p), \quad p \in \{1, ..., K\}
\end{equation}
where $\mathcal{F}_{\text{Director}}(\cdot)$ is the decomposition function that segments the script $S$ into meaningful sub-units $\mathcal{S}$ based on character interactions, thematic continuity, and narrative flow.
Specifically, the director agent follows a structured reasoning process:
\textit{1) Identify Core Narrative Structure}: The director agent first analyzes the synopsis to identify main acts, key plot points, and turning points.
\textit{2) Define Script Segmentation}: Based on these core narrative elements, the script synopsis is divided into discrete, self-contained sub-scripts (\( S_p \)).
\textit{3) Ensure Logical Story Progression}: Each sub-script \( S_p \) maintains temporal and thematic coherence across \( \mathcal{S} \) for a cohesive plot.
\textit{4) Maintain Character Consistency}: The segmentation preserves the roles and relationships of characters from set \( C \), ensuring their presence and interactions remain accurate throughout \( \mathcal{S} \).
\textit{5) Justify the Division}: For each sub-script, a clear rationale for its segmentation (\eg{} major event shift, emotional climax, new setting introduction) must be provided, serving as a reference for the subsequent step-by-step reasoning process.

%

\subsubsection{Scene Plan Agent}
\label{Scene}
With sub-scripts $\mathcal{S}$, the next step involves determining the movie scenes, key scene elements, and scene boundaries.
The Scene Plan Agent is designed to refine sub-scripts $\mathcal{S}$ into scene compositions $\mathcal{P} = \{P_1, P_2, ..., P_N\}$, where each $P_i$ represents a detailed scene with enriched descriptions for the $i$-th scene.
The scene planning process is formalized as:
\begin{equation}
    \mathcal{P} = \sum_{p=1}^{K} \mathcal{F}_{\text{Scene}}(S_p, C, i), \quad i \in \{1, \dots, N\}
\end{equation}
where $\mathcal{F}_{\text{Scene}}(\cdot)$ denotes the scene plan agent, which outputs a refined scene list $\mathcal{P}$. 
For the $i$-th scene $P_i$,
the scene plan agent comprehensively summarizes factors such as involved characters, plot, emotional tone, visual style, and cinematography notes to thoroughly define the scene variables and expressive elements.
Similar to the director agent, the scene plan agent follow a structured reasoning process: 
\textit{1) Analyze the Narrative Structure}: The agent identifies key turning points and transitions, ensuring each scene forms a complete narrative with a clear start and end.
\textit{2) Extract Key Scene Elements}: The model identifies all characters, their roles, interactions, and key events in each major scene.
\textit{3) Define Scene Boundaries}: Finally, identify natural story breaks (\eg{} location shifts, time jumps, emotional climaxes), ensuring each scene has a clear purpose, and justify each division (\eg{} tone shift, new conflict).
\textit{4) Justify the Division}: Preserve the internal Chain-of-Thought behind scene segmentation and reasoning to ensure traceability and analyzability.

\subsubsection{Shot Plan Agent}
\label{Shot}
%
Given the structured scenes $\mathcal{P}$, the shot plan agent is responsible for defining shot-level details, including character-aware plot, cinematographic parameters, and visual dynamics. 
Specifically, each scene $P_i$ is further decomposed into detailed shot compositions $\mathcal{V}^i = \{V^i_1, V^i_2, ..., V^i_M\}$, where each shot $V_j^i$ captures distinct visual perspectives and cinematographic intentions of the $j$-th shot video in the $i$-th scene.
Therefore, with scene list $\mathcal{P} = \{P_1, P_2, ..., P_N\}$, the formalized function for shot-level decomposition is expressed as:
\begin{equation}
    \mathcal{V} = \sum_{i=1}^{N} \mathcal{F}_{\text{Shot}}(P^i, C, j), \quad j \in \{1, \dots, M\}
\end{equation}
where $\mathcal{F}_{\text{Shot}}(\cdot)$ is the Shot Plan Agent function responsible for generating structured shots $\mathcal{V} = \{ V_j^i \mid i = 1, 2, \dots, N, \; j = 1, 2, \dots, M \}$.
Each shot level script $V_j^i$ includes rich, structured shot script annotation, such as the involved characters, plot, camera movements, shot type, and character subtitles.

The Shot Plan Agent follows a structured reasoning workflow:
\textit{1) Determine Shot Composition and Framing}: Identify appropriate shot types (\eg{} wide, medium, close-up) and camera angles based on scene content and emotional impact.
\textit{2) Specify Cinematographic Techniques}: Clearly define camera movements (static, pan, tilt, zoom, tracking), lighting styles, and visual effects necessary.
\textit{3) Coordinate Visual Continuity}: Ensure visual coherence and consistency across shots within each scene, avoiding abrupt transitions or inconsistent visual styles.
\textit{4) Align with Scene Narrative}: Each shot should advance the narrative or enhance emotions, with clear reasoning for traceability and analysis.

\begin{figure}[t]
	\includegraphics[width=0.99\linewidth]{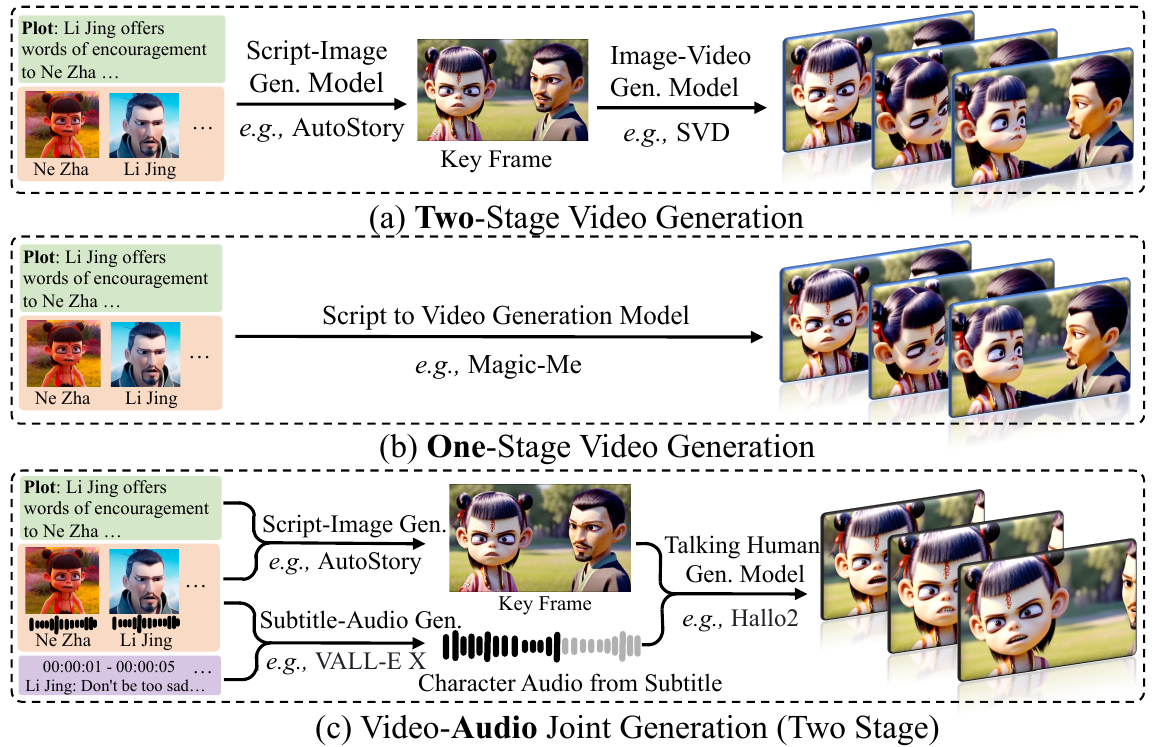}
	\vspace{-0.1cm}
	\caption{\textbf{Customized Shot-Level Video Generation for \Ours.} 
    Current shot-level character-aware video generation approaches can be divided into three categories: (a) Keyframe-based two-stage video generation; (b) One-stage end-to-end video generation; (c) Keyframe-based joint video and audio generation.
    }
\label{videgen}
\end{figure}

\subsubsection{Customized Video and Audio Generation}
\label{videogen}
At this stage, given the shot level script annotation $V_{j}^i$ for the $j$-th shot in the $i$-th scene, the model invokes various customized image and video generation models~(\eg{} AutoStory~\cite{wang2024autostory}, StoryDiffusion~\cite{zhou2024storydiffusion}, and Magic-Me~\cite{ma2024magic}) are used to produce the final shot-level video $\widehat{V}_{j}^i$.

Current video generation models, such as SVD~\cite{svd}, DreamVideo~\cite{wei2024dreamvideo} are unable to simultaneously support subtitle-to-audio generation.
For the talking human generation task, some priors, such as Hallo2~\cite{cui2024hallo2} and Edtalk~\cite{tan2024edtalk} primarily focuses on single human generation.
%
%
Therefore, our current technology cannot fully address the simultaneous audio-video generation in a single model.
Based on whether audio generation is required, we categorize the movie generation setting into two task:
\begin{itemize}[leftmargin=*]
    \item \textbf{Pure Shot-level Video Generation in Figure~\ref{videgen} (a)-(b).}
    In this setting, we do not consider the audio generation of subtitle for the characters.
    Instead, we focus solely on generating pure video by modeling:
    $\widehat{\mathcal{V}}_j^i = \mathcal{F}_{\text{Video}}(V_{j}^i, C)$, where $\mathcal{F}_{\text{Video}}(\cdot)$ can either be a two-stage video generation model~(\eg{} the combination of StoryDiffusion~\cite{zhou2024storydiffusion} and CogVideoX~\cite{yang2024cogvideox}) or a one-stage customized end-to-end video generation model~(\eg{} Magic-Me~\cite{ma2024magic}), as illustrated in Figure~\ref{videgen} (a)-(b).

    \item \textbf{Video and Audio Joint Generation in Figure~\ref{videgen} (c).}
    %
    %
    In this setting, the character bank $C$ must include the voice sample of each character, represented as $\{[\texttt{char}_k, I_k, A_k]\}_{k=1}^{L}$. 
    Since no current model can simultaneously generate both audio and video, we adopt a two-stage video-audio joint generation strategy, as shown in Figure~\ref{videgen} (c).
    Formally, we express the formulation as:
    $\widehat{V}_j^i = \mathcal{F}_{\text{Talking}}\left(\mathcal{F}_{\text{Image}}(V_{j}^i, C),\, \mathcal{F}_{\text{Audio}}(V_{j}^i, C)\right)$,
    where $V_{j}^i$ includes subtitles for all characters at the shot level. 
    And $\mathcal{F}_{\text{Talking}}(\cdot)$, $\mathcal{F}_{\text{Image}}(\cdot)$, and $\mathcal{F}_{\text{Audio}}(\cdot)$ denote the talking-human generation model~(\eg{} Hallo2~\cite{cui2024hallo2}), customized image generation model~(\eg{} StoryDiffusion~\cite{zhou2024storydiffusion}), and customized audio generation model~(\eg{} VALL-E X~\cite{zhang2023speak}), respectively.

\end{itemize}

\begin{figure}[t]
	\includegraphics[width=0.99\linewidth]{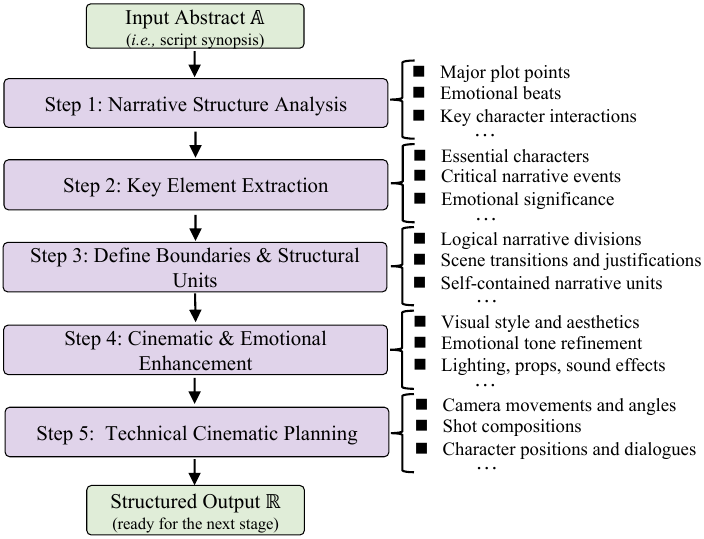}
	\caption{\textbf{Flowchart of the Internal Chain-of-Thought reasoning process.}
    Through Internal CoT, various agents can process and manage narrative elements more efficiently.
    }
\label{fig:internal_cot_flowchart}
\end{figure}

\begin{figure*}[t]
	\includegraphics[width=0.99\linewidth]{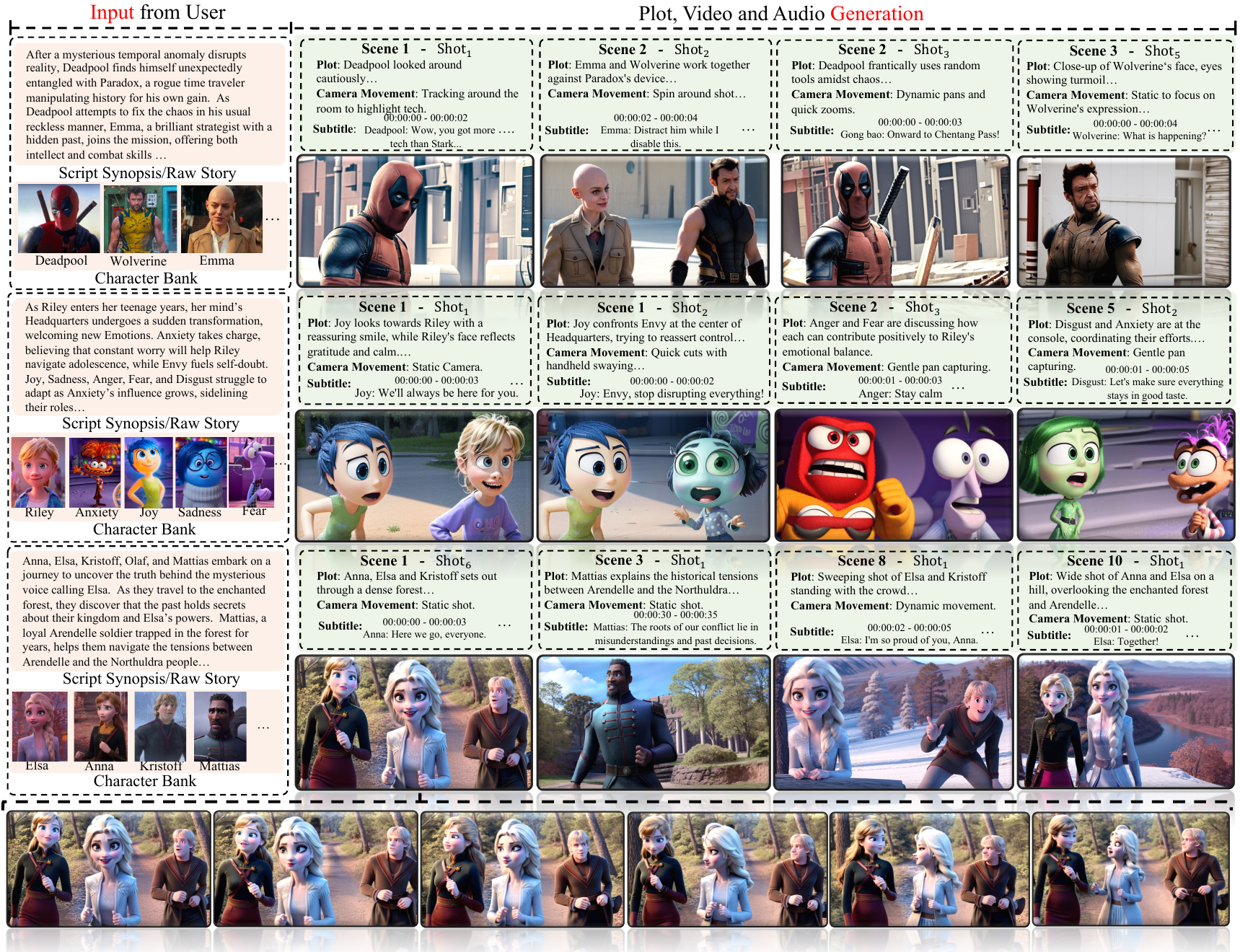}
	\vspace{-0.4cm}
	\caption{\textbf{More Visualizations for \Ours.} 
     Our \Ours can generate coherent storylines and detailed shots.
    }
    \vspace{-0.3cm}
\label{visualization2}
\end{figure*}

\subsection{Internal Chain of Thought}
\label{cottt}
The Internal Chain-of-Thought  provides a general structured reasoning framework employed by various planning agents (\textit{e.g.,} Scene Plan Agent, Shot Plan Agent, Director Agent) to methodically translate abstract narrative and cinematic requirements into detailed, actionable plans.
Formally, given an input abstract $\mathbb{A}$ (\textit{e.g.,} scene synopsis, script synopsis), the Internal CoT generates structured reasoning:
$\mathbb{R} = \mathcal{F}_{\text{CoT}}(\mathbb{A}),
$
where $\mathcal{F}_{\text{CoT}}(\cdot)$ denotes the general internal reasoning function utilized by planning agents, involving explicit, systematic reasoning steps prior to the final generation of detailed cinematic output.

As shown in Figure~\ref{fig:internal_cot_flowchart}, the Internal CoT generally encompasses the five stages:
(1) Narrative Structure Analysis, which identifies major plot points, emotional beats, and key character interactions; 
(2) Key Element Extraction, focusing on essential characters, critical narrative events, and emotional significance; 
(3) Define Boundaries \& Structural Units, establishing logical narrative divisions, scene transitions, and self-contained units; 
(4) Cinematic \& Emotional Enhancement, refining visual style, emotional tone, lighting, props, and sound effects; 
and (5) Technical Cinematic Planning, specifying camera movements, shot compositions, character positions, and dialogues. 
Upon completion of this structured internal reasoning, agents produce a comprehensive, structured output that encapsulates all detailed planning elements necessary for subsequent execution by next stage.

\section{Experiments}
\label{sec:Experiments}

\begin{table*}[t]
    \centering
    \small 
    \setlength{\tabcolsep}{1mm}
    \caption{\textbf{Performance of Automatic metric for Script to Keyframe/Video Generation on \Ours.} 
    Models without character consistency~(\eg Open-Sora~\cite{opensora}) are excluded.
    `Subject Cons.', `Bg Cons.', `Motion Smth.', and `Dyn. Degree' refer to `Subject Consistency', `Background Consistency', `Motion Smoothness', and `Dynamic Degree' from the advanced VBench Metrics~\cite{vbench}, respectively.
    GPT-4o serves as the LLM. And \Ours adopts multi-agent and internal CoT reasoning, while others rely on single-step generation.
    }
\begin{tabular}{c|cc|ccccc}
    \multirow{2}{*}{Method}   & \multirow{2}{*}{CLIP$ \uparrow $} & \multirow{2}{*}{Inception$ \uparrow $}   & \multicolumn{5}{c}{VBench Metircs~\cite{huang2024vbench}/\% $\uparrow$ }   \\
    \cline{4-8} 
     &  & &  Subject Cons. & Bg Cons. & Motion Smth.& Dyn. Degree & Aesthetic
    \\
    \hline
    \multicolumn{8}{l}{\textit{Script Synopsis to Keyframe/StoryBoard Generation}}  \\
    \hline
    
    
    StoryGen~\cite{liu2024intelligent} &  19.73  &  6.21    & - & - & -& - & - \\
    
    StoryDiffusion~\cite{zhou2024storydiffusion} &  20.46  &  6.24    & - & - & -& - & - \\

    AutoStory~\cite{wang2024autostory}  & 20.21 &  6.01  & - & - & -& - & - \\

    \Ours  & \textbf{22.12}   &  \textbf{7.23}  & - & - & -& - & - \\
    \hline

    \multicolumn{8}{l}{\textit{Script Synopsis to Movie Generation}}  \\
    \hline

    StoryDiffusion~\cite{zhou2024storydiffusion} + SVD~\cite{svd} & 21.39 &    8.36   & 93.64  & 93.78  & 96.30 & 74.48 & 56.69 \\
    
    
    StoryDiffusion~\cite{zhou2024storydiffusion} + CogVideoX~\cite{yang2024cogvideox} &   21.83 &  9.01 &   93.45 & 94.56 &  96.60  &  27.89 & 56.05
     \\

    AutoStory~\cite{wang2024autostory} + CogVideoX~\cite{yang2024cogvideox} &   20.27 & 7.21  &  91.45 & 93.32 & 95.87   & 70.32 & 52.34
     \\

    DreamVideo~\cite{wei2024dreamvideo} &  21.37  & 8.11  &   93.17 &  93.77&  96.40  &  26.97 & 42.16
     \\

    Magic-Me ~\cite{ma2024magic} &  21.72  & 8.34   & 94.01 & \textbf{94.68} &  96.41  &  14.86 & 55.89
    \\
     
    \Ours  &  \textbf{22.25}  &  \textbf{9.39}  & \textbf{94.72} & 93.52 & \textbf{97.84} & \textbf{76.27}  & \textbf{58.63}
\end{tabular}

    \label{tab:experiment}
\end{table*}

\subsection{Experiment Setting}

\textbf{Metric.} Following prior works~\cite{zheng2024videogen,chen2023pixart}, we evaluate the model using both automated metrics~(\eg{} VBench~\cite{vbench}) and human voting. 
Automated metrics offer objective analysis but may not fully match human preferences, while user studies capture real preferences but can be biased.
Since automated movie generation varies in shot count and lacks a ground truth video, metrics like FID~\cite{heusel2017gans} cannot be computed.
Two A6000 GPUs were used for all experiments.

\textbf{Baseline.} 
%
Existing approaches, such as DreamFactory~\cite{xie2024dreamfactory}, StoryAgent~\cite{hu2024storyagent}, and StoryDiffusion~\cite{zhou2024storydiffusion}, fail to address the automated movie generation task~(Section~\ref{TaskFormulation}), struggling with multi-characters consistency and automated script planning.
Therefore, we decompose the task into three components: LLM-based script processing~(GPT4-o~\cite{openai2025gpt4o}, Deepseek-R1~\cite{guo2025deepseek}, Llama3.3~\cite{dubey2024llama}), image generation~(AutoStory~\cite{wang2024autostory},StoryDiffusion~\cite{zhou2024storydiffusion}), and video generation~(DreamVideo~\cite{wei2024dreamvideo},Magic-Me~\cite{ma2024magic}). 
For each component, we incorporate baseline models for evaluation and comparison, as detailed in Table~\ref{tab:experiment}.

\textbf{Evaluation Dataset.} Since the automated movie generation task (Section~\ref{TaskFormulation}) is formally defined for the first time, we need to construct a new evaluation dataset.
This evaluation dataset takes as input a script summary, character names, photos, and audio samples, and outputs a series of shot videos.
To achieve this, we propose a test set, namely MoviePrompts, consisting of $10$ script prompts: $8$ prompts are derived from well-known movies~(\eg{} Ne Zha 2, Frozen II, and Inside Out 2), while the remaining $2$ prompts~(\eg{} Fictional stories and characters) are privately designed by two annotators.
%

\subsection{Performance Comparisons and Analysis}

\subsubsection{Automatic Metric on MoviePrompts}
Table~\ref{tab:experiment} experimental results for script-to-keyframe and script-to-movie generation.
In keyframe generation, \Ours achieves the highest CLIP score $22.12$ and Inception scores $7.23$, indicating superior visual-semantic alignment and image quality.
For movie generation, \Ours consistently outperforms nearly all baselines across VBench metrics~\cite{vbench}, achieving the highest motion smoothness $97.84$, dynamic degree $76.27$, and aesthetic quality $58.63$.
These results highlight \Ours as a new state-of-the-art for automatic story-based video generation.

\begin{figure}[t]
	\includegraphics[width=0.99\linewidth]{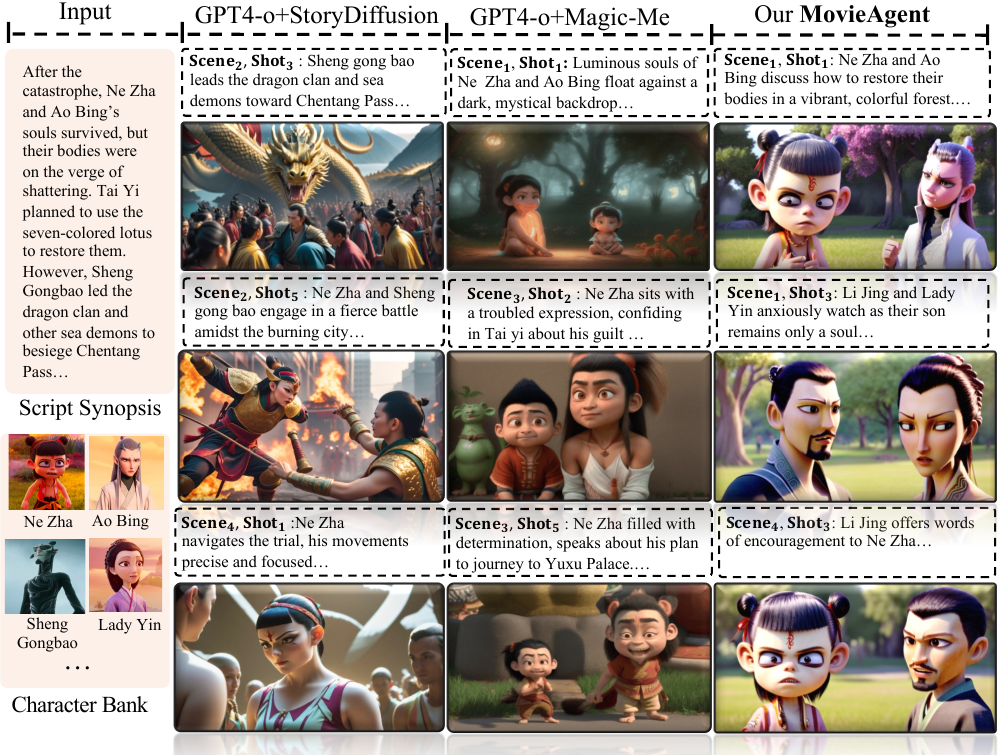}
	\vspace{-0.2cm}
	\caption{\textbf{ Visualization Comparison for Different Methods.} 
    }
    \vspace{-0.1cm}
\label{visualization_comparison}
\end{figure}

\subsubsection{Human Rating}
\begin{figure*}[t]
	\includegraphics[width=0.99\linewidth]{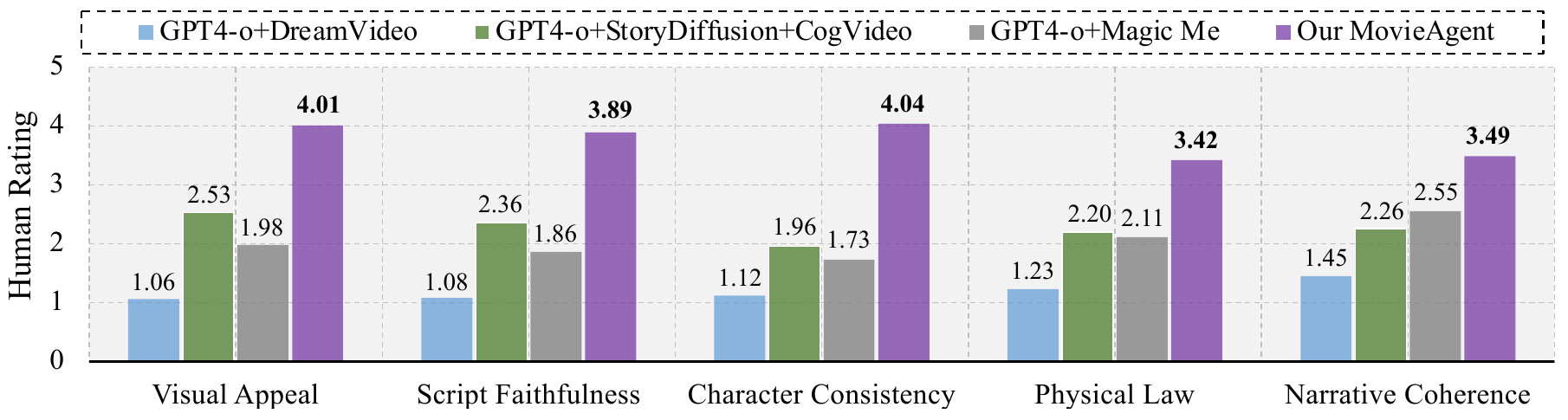}
	\vspace{-0.2cm}
	\caption{\textbf{Performance of User Study for Automated Movie Generation on \Ours.} 
     Models without character consistency~(\eg Open-Sora~\cite{opensora}) are excluded.
     Under the $0$-$5$ score rating system,
     \Ours demonstrates outstanding performance across multiple aspects, particularly in script faithfulness and character consistency.
    }
    \vspace{-0.1cm}
\label{UserStudy}
\end{figure*}

Figure~\ref{UserStudy} presents the user study for \Ours from two expert evaluators on the MoviePrompts dataset ($10$ movies).
For a fairer and more fine-grained comparison, evaluators need rate each shot video on a scale of $1$ to $5$ based on the corresponding evaluation rules (detailed rules see supplementary materials).
Due to the high cost of human ratings, we limited assessments to key baselines (GPT-4o with DreamVideo and Magic Me).
\Ours present a promising performance, outperforming the best baseline by up to $2$ points on a five-point scale.
Notably, it excels in Narrative Coherence ($3.49$), Visual Appeal ($4.01$), Script Faithfulness ($3.89$), Character Consistency ($4.04$), and Physical Law ($3.42$).
Figure~\ref{visualization_comparison} provides a relevant visual comparison, while Figure~\ref{visualization2} presents additional visualizations, including coherent storylines, keyframes, and shot videos.
These results highlight the effectiveness of our multi-agent and CoT reasoning approach.

\begin{table*}[t]
    \centering
    \small 
    \setlength{\tabcolsep}{1mm}
    \caption{\textbf{Ablation Study for LLM, CoT, and Multi-Agent.} 
    In this experiment, ROICtrl~\cite{gu2024roictrl} and CogVideoX~\cite{yang2024cogvideox} are used as the image and video generation methods.
    Due to the high cost of human evaluation (up to $100$ shots per movie), our ablation study focuses on three movies: Ne Zha 2, Frozen II, and Inside Out 2.
    }
\begin{tabular}{c|cc|c@{\hskip 10pt}c@{\hskip 10pt}c@{\hskip 10pt}c@{\hskip 10pt}c|c}
    LLM Model  & Internal CoT & Multi-Agent & Vis. Appeal & Script Faith. &
    Char. Consist. & Phys. Law  & Narr. Coher. &
   Average\\
    \hline

    Llama3.3-70b & \ding{51} & \ding{51} & 3.86 & 3.50  & 4.00 & 2.70 & 3.13 & 3.44 \\
    
    Deepseek-V3 & \ding{51} & \ding{51} & 3.74 & 3.66  & 3.78 &  3.07 & 3.45 &  3.54  \\
    
    Deepseek-R1 & \ding{51}&  \ding{51} & 4.02 &  3.59 & 4.09  & 3.38 & \textbf{3.79}  & 3.78   \\
    

    \hline
    GPT4-o &  &  & 3.89 & 3.36 & 4.08&  3.37  & 3.09  & 3.55 
    \\
    
    GPT4-o &  & \ding{51} & 4.02& 3.69  &  4.13  & 3.46  & 3.31 & 3.72 
    \\
    
    GPT4-o & \ding{51} & & 3.92 & 3.38 & 4.08  &  3.37  & 3.29  & 3.61 
    
    \\
    GPT4-o & \ding{51} & \ding{51} & \textbf{4.04} & \textbf{3.92}  & \textbf{4.11}   & \textbf{3.49}  & 3.55 & \textbf{3.82}
\end{tabular}
    \label{tab:ablation_llm}
\end{table*}

\subsection{Ablation Study}

We ablated three key aspects of \Ours: Internal CoT, LLM, and Multi-Agent.
%
%
Currently, there are no automated metrics for evaluating script faithfulness and narrative coherence, which are key aspects of movie script and quality generation.
Moreover, priors~\cite{podell2023sdxl, chen2023pixart,wu2023paragraph} confirm that human evaluation is more reliable than automated metrics for generation tasks.
Therefore, our study focuses on human assessment.

\subsubsection{Effect of Internal Chain of Thought}
Table~\ref{tab:ablation_llm} presents the ablation study for the internal Chain of Thought.
Results show that GPT-4o with Internal CoT achieves a slight average score improvement ($3.61$ vs. $3.55$ without CoT), with notable gains in Narrative Coherence ($3.29$ $vs.$ $3.09$).
This is expected, as incorporating CoT enables step-by-step reasoning during story script generation, enhancing logical flow and coherence.
By breaking down the reasoning process, CoT helps maintain narrative consistency and structure, as illustrated in Figure~\ref{fig:internal_cot_flowchart}.

\subsubsection{Effect of Large Language Model}
Table~\ref{tab:ablation_llm}  compares the performance of various LLMs: Llama3.3-70b, Deepseek-V3, Deepseek-R1, and GPT4-o across human evaluation metrics including visual appeal, script faithfulness, character consistency, physical law, narrative coherence, and average.
The results reveal that GPT4-o, particularly with multi-agent collaboration, achieves the highest average score of $3.82$, outperforming Deepseek-V3 $3.54$, and Deepseek-R1 $3.78$. 
However, GPT-4o underperforms Deepseek-R1 in Narrative Coherence ($3.55$ vs. $3.79$). 
This is expected, as Deepseek-R1 is optimized for reasoning tasks with a built-in Internal CoT process, enabling more extensive reasoning and generating smoother, richer movie narratives.

\subsubsection{Effect of Multi-Agent}
Table~\ref{tab:ablation_llm} presents the evaluation for the impact of multi-agent collaboration.
The results show that multi-agent collaboration significantly enhances performance, with GPT4-o achieving an average score of $3.72$, compared to $3.55$ without multi-agent collaboration.
The improvements in script faithfulness and narrative coherence are particularly significant, with increases of $0.33$ and $0.22$ on a five-point scale, respectively.
This is reasonable because multi-agent system is a hierarchical structure. 
Multi-agent collaboration is more efficient than single-step script generation, as it better translates a script synopsis into a full movie script with detailed plot logic and parameters at the scene and shot levels.

\section{Conclusion}

In this paper, we firstly explore and define the paradigm of automated movie/long-video generation and propose \Ours, a multi-agent CoT-based framework for automated filmmaking.
By integrating hierarchical reasoning and specialized AI agents, \Ours automates story structuring, scene planning, and shot composition, reducing human intervention while ensuring narrative coherence and cinematographic quality.
Experiments show that \Ours improves story consistency, character preservation, and audiovisual synchronization, addressing key challenges in AI-driven filmmaking.
Our approach provides a scalable solution for automated storytelling, offering new insights into the future of AI-assisted movie production.

\section{Appendix}

\begin{table}[t]
    \centering
    \small 
    \setlength{\tabcolsep}{1mm}
    \caption{\textbf{Metric Summary.} 
    We evaluate the generated long video on a 0-5 scale across five key aspects.
    }
    
\begin{tabular}{p{2.8cm}|p{5cm}}
    Metric & Description \\
    \hline
    Visual Appeal & Evaluates the overall visual quality, realism, and aesthetic consistency of the generated video. \\
    Script Faithfulness & Measures how accurately the generated shot-level video content follows the provided shot level script and storyline. \\
    Narrative Coherence & Assesses whether the narrative flows logically, maintaining consistent plot development. \\
    Character Consistency & Checks whether characters maintain a stable appearance, behavior, and role throughout the movie. \\
    Physical Law & Evaluates whether the generated video adheres to basic physical laws (\eg{} gravity, motion realism). \\
\end{tabular}

    \label{tab:metric}
\end{table}

\begin{figure}[t]
    \centering
	\includegraphics[width=0.95\linewidth]{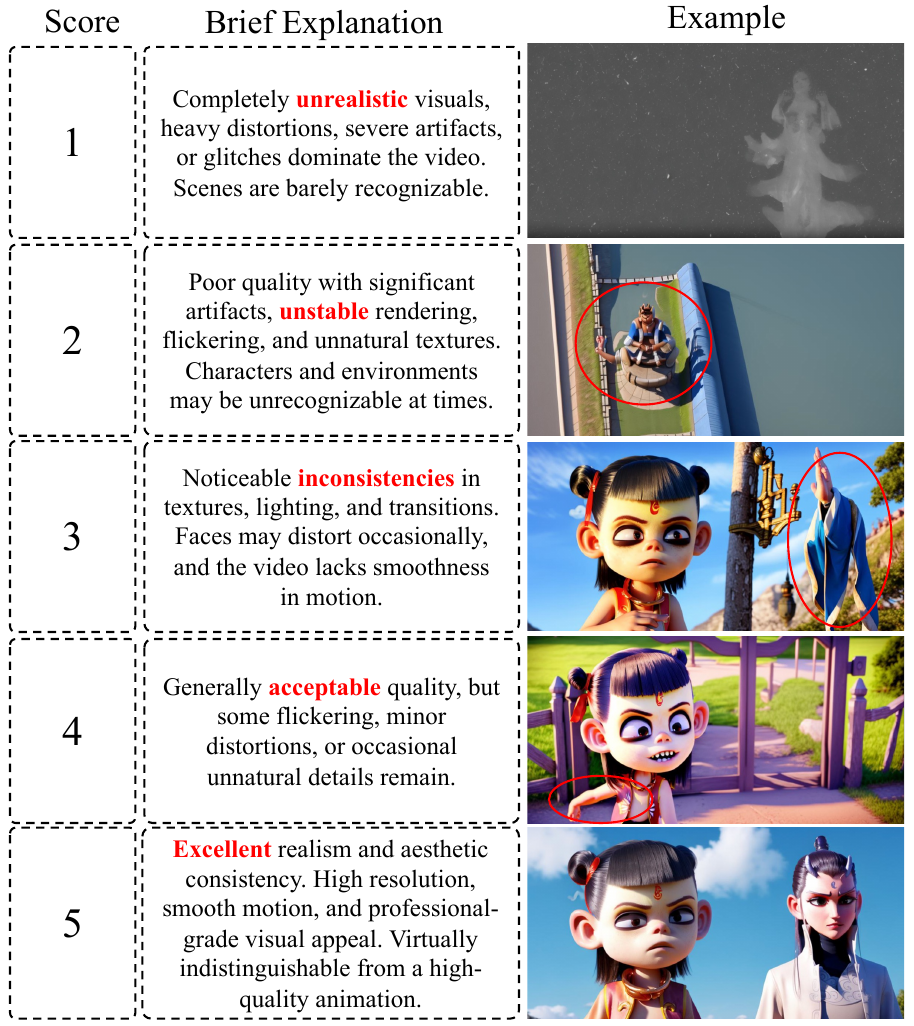}
	\vspace{-0.2cm}
	\caption{\textbf{Metric Criteria for Visual Appeal.} 
    }
    \vspace{-0.1cm}
\label{user_study_metric_Visual}
\end{figure}

\begin{figure}[t]
    \centering
	\includegraphics[width=0.95\linewidth]{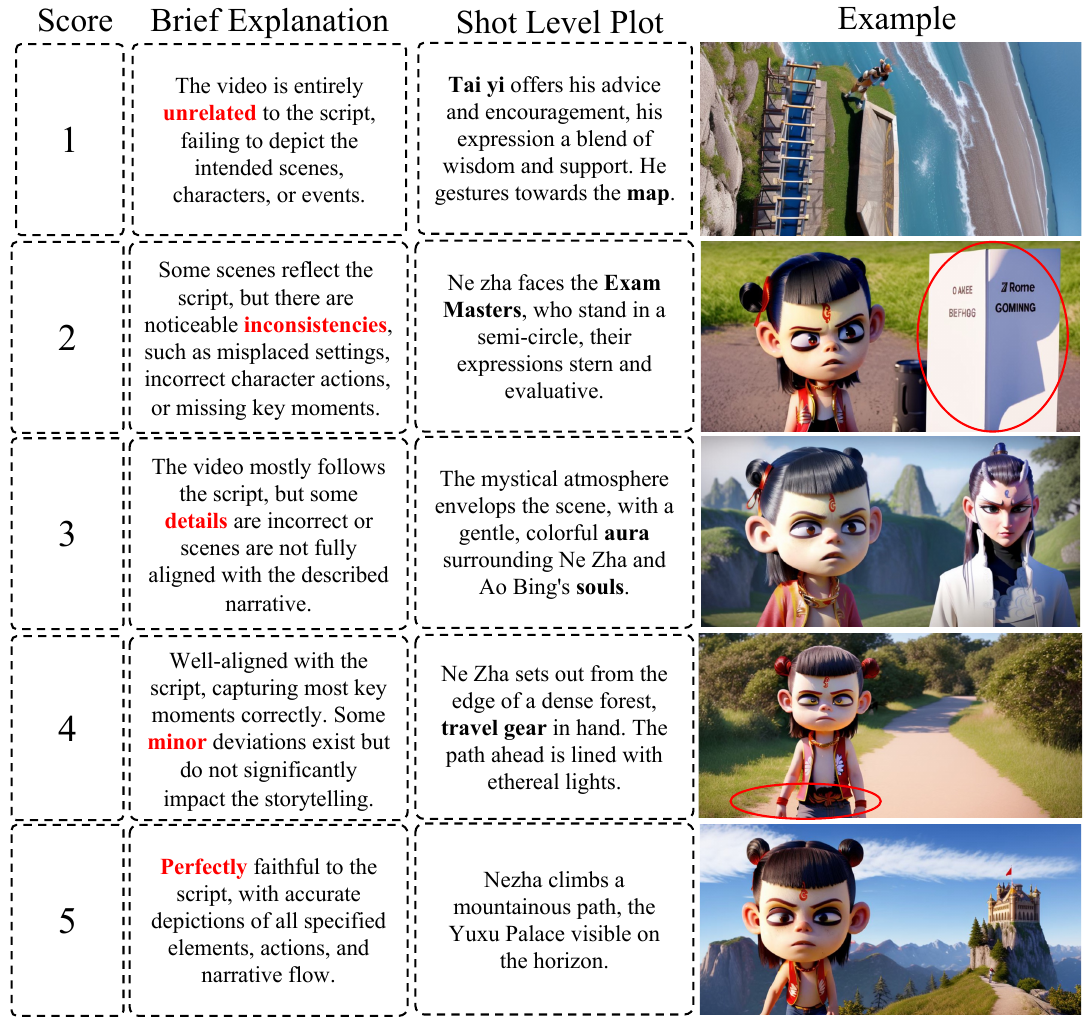}
	\vspace{-0.2cm}
	\caption{\textbf{Metric Criteria for Script Faithfulness.} 
    }
    \vspace{-0.1cm}
\label{user_study_metric_ScriptFaithfulness}
\end{figure}

\begin{figure*}[t]
    \centering
	\includegraphics[width=0.90\linewidth]{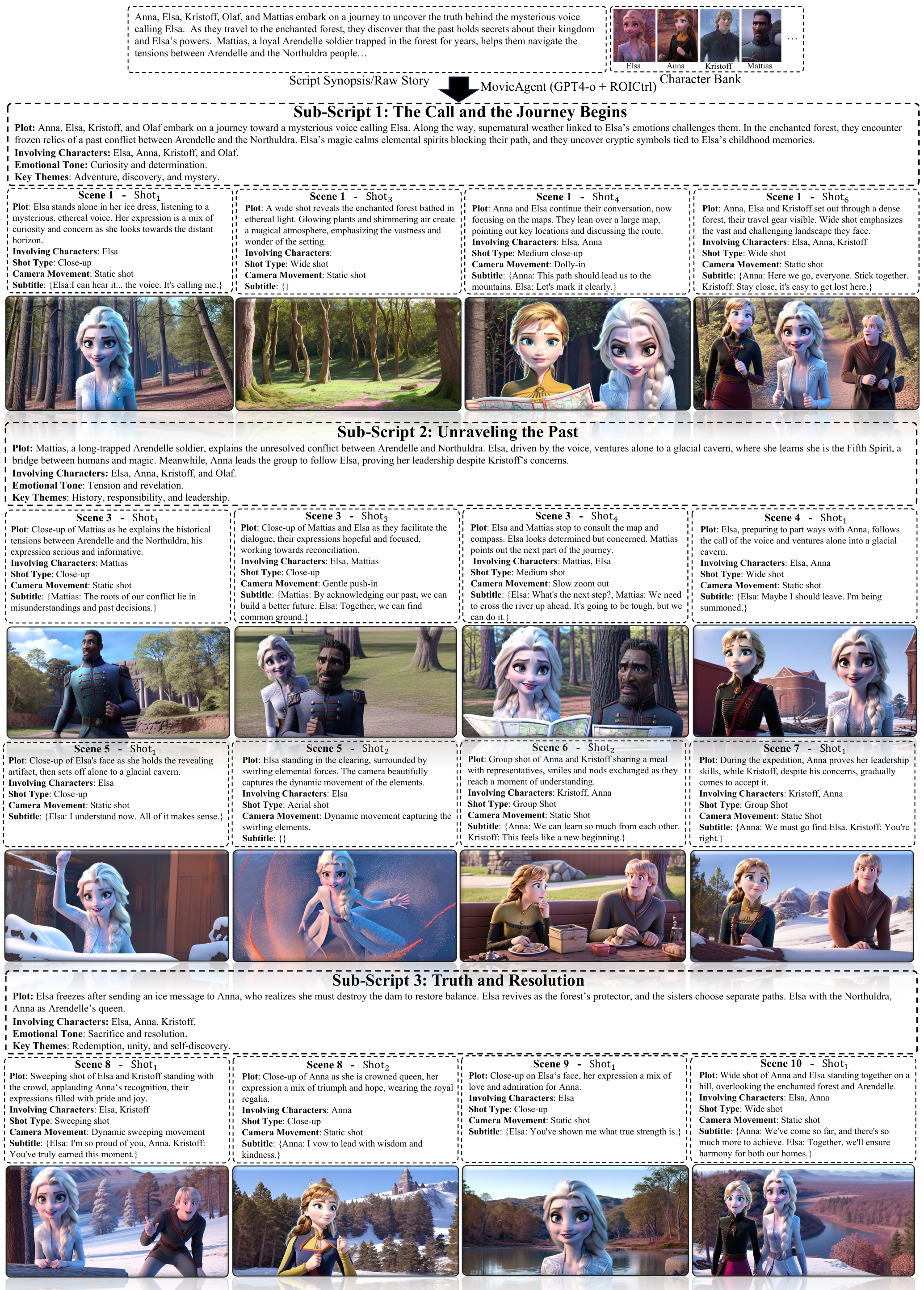}
	\vspace{-0.3cm}
	\caption{\textbf{Detailed Visualization for Movie~(\eg{} FrozenII) Generation from \Ours.} 
    }
    \vspace{-0.1cm}
\label{more_visualization}
\end{figure*}

\section{Metrics of Human Evaluation}
\subsection{Definition for Metrics}
To systematically evaluate the quality of generated long videos, we assess five key aspects on a $0$-$5$ scale: visual appeal, script faithfulness, narrative coherence, character consistency, and adherence to physical laws.
These metrics collectively measure both the aesthetic and structural integrity of the video. 
Specifically, they ensure that the visual quality remains realistic and consistent, the generated content accurately follows the script and storyline, and the narrative flows logically without abrupt transitions. 
Additionally, character consistency is crucial for maintaining a coherent identity and behavior throughout the video, while adherence to physical laws enhances motion realism and natural interactions within the scene. 
Together, these criteria provide a comprehensive framework for evaluating the effectiveness and realism of long video generation.

\subsection{Rules for User Study}
In this user study, annotators are required to rate each generated long video on a $0$-$5$ scale across five key evaluation aspects: Visual Appeal, Script Faithfulness, Narrative Coherence, Character Consistency, and Physical Law Adherence. 
Before scoring, the detailed criteria for each metric are established and provided as guidelines to ensure consistent evaluation.

\begin{figure*}[t]
    \centering
	\includegraphics[width=0.98\linewidth]{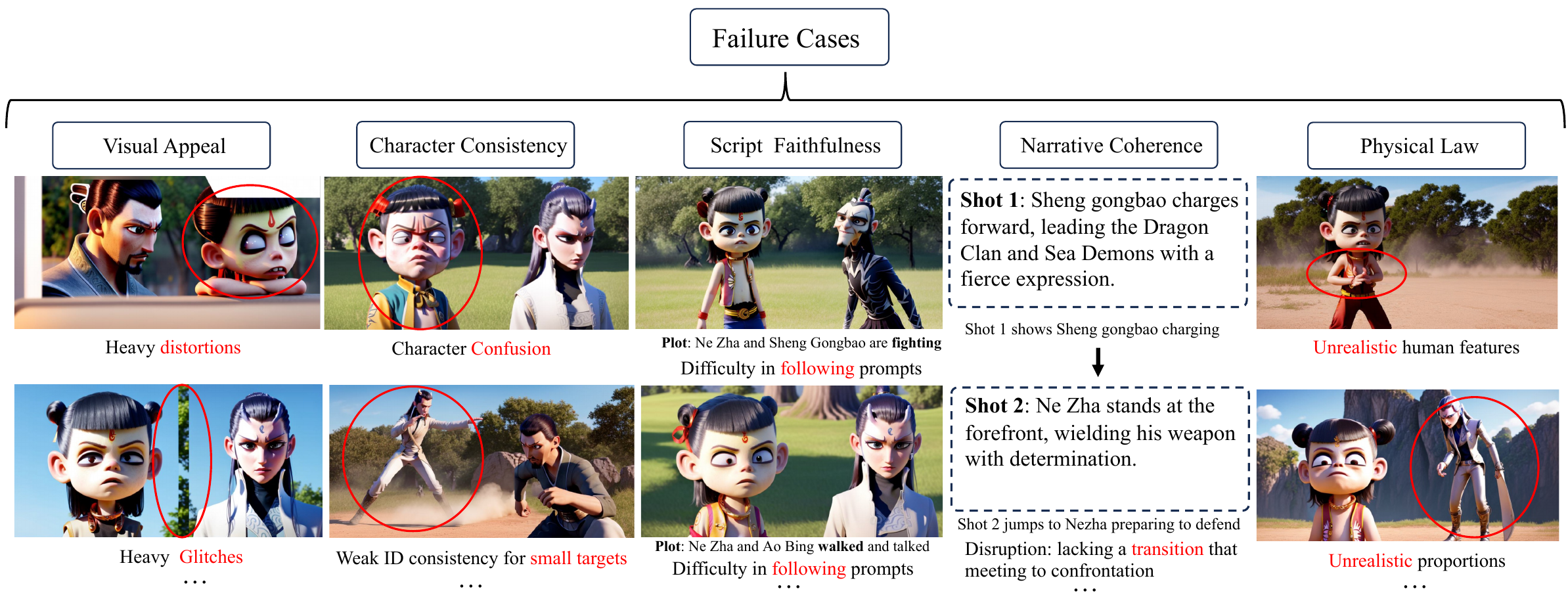}
	\vspace{-0.3cm}
	\caption{\textbf{Failure Cases for Movie Generation from \Ours.} 
    }
    \vspace{-0.1cm}
\label{badcase}
\end{figure*}

\textbf{Visual Appeal.} The metric evaluates key factors such as realism, aesthetic consistency, artifact presence, motion smoothness, and rendering quality.
Figure~\ref{user_study_metric_Visual} presents the detailed criteria and examples corresponding to each score.
\begin{itemize}
    \item \textit{Score 1 (Severe Artifacts):} Heavy distortions, glitches, and unrecognizable scenes dominate the video.
    \item \textit{Score 2 (Poor Quality):} Strong flickering, unstable textures, and character inconsistencies.
    \item \textit{Score 3 (Inconsistent Rendering):} Noticeable issues in textures, lighting, or motion transitions, though the content remains understandable.
    \item \textit{Score 4 (Acceptable Quality):} Good visual appeal with minor distortions or flickering but no major distractions.
    \item \textit{Score 5 (High-Quality Rendering):} Smooth, realistic, and professionally rendered visuals with high aesthetic consistency.
\end{itemize}
By applying this scoring framework, human evaluators can systematically assess the visual quality of generated videos, ensuring a consistent and objective evaluation standard.

\textbf{Script Faithfulness.} %
The metric assesses whether the generated content correctly follows the described plot, character actions, scene settings, and overall narrative structure. Figure~\ref{user_study_metric_ScriptFaithfulness} presents the detailed criteria.
\begin{itemize}
    \item \textit{Score 1 (Unrelated to Script):} The video fails to depict the intended scenes, characters, or events, making it entirely disconnected from the script.
    \item \textit{Score 2 (Major Inconsistencies):} Some script elements are present, but there are misplaced settings, incorrect character actions, or missing key moments.
    \item \textit{Score 3 (Partial Alignment):} The video follows the script’s general structure but contains minor inaccuracies in details, scene transitions, or character behavior.
    \item \textit{Score 4 (Good Faithfulness):} Most key moments are accurately captured with only minor deviations that do not significantly affect storytelling.
    \item \textit{Score 5 (Perfect Script Adherence):} The video precisely reflects all specified elements, actions, and plot progression, achieving full alignment with the script.
\end{itemize}
This structured scoring system ensures an objective and consistent evaluation of script adherence in generated videos, helping assess their narrative integrity.

\textbf{Narrative Coherence.} 
Narrative coherence assesses whether the storyline flows logically, maintaining consistent plot progression and smooth scene transitions. 
The detailed scoring criteria are as follows:
\begin{itemize}
    \item \textit{Score 1 (Completely Incoherent):} The video lacks any meaningful structure, consisting of random and disconnected scenes that fail to form a logical storyline.
    \item \textit{Score 2 (Frequent Disruptions):} Contains abrupt cuts, illogical progressions, and a lack of continuity, making the plot difficult to follow.
    \item \textit{Score 3 (Disjointed Flow):} Some attempt at a storyline exists, but inconsistencies in sequencing and unnatural pacing disrupt the narrative flow.
    \item \textit{Score 4 (Well-Structured Narrative):} The storyline is logically structured with smooth transitions, though minor pacing issues may still be present.
    \item \textit{Score 5 (Fully Coherent and Engaging):} The video exhibits strong storytelling, seamless scene transitions, and consistent plot development, ensuring an immersive narrative experience.
\end{itemize}

\textbf{Character Consistency.} 
Character consistency evaluates whether characters maintain a stable appearance, behavior, and role throughout the video.
The detailed scoring criteria are as follows:
\begin{itemize}
    \item \textit{Score 1 (Completely Inconsistent):} Character appearances change frequently, making them unrecognizable. 
    \item \textit{Score 2 (Severe Inconsistencies):} Major discrepancies in facial features, outfits, and personality.
    \item \textit{Score 3 (Noticeable Variations):} Some variations in facial structure, clothing, or expressions exist, and behaviors may not always align with character roles.
    \item \textit{Score 4 (Good Consistency):} Appearance and behavior remain stable, with only minor variations that do not significantly impact immersion.
    \item \textit{Score 5 (Perfect Consistency):} Characters maintain a stable identity, facial features, and clothing consistently throughout the entire video.
\end{itemize}

\textbf{Physical Law.} 
Physical law adherence assesses whether the video follows basic physics principles, including motion realism, object interactions, and environmental consistency. 
The detailed scoring criteria are as follows:
\begin{itemize}
    \item \textit{Score 1 (Completely Unrealistic Physics):} Objects float, characters phase through walls, and movements defy gravity without reason.
    \item \textit{Score 2 (Frequent Violations):} Major inconsistencies, such as erratic character movements, unnatural collisions, or unrealistic gravity interactions.
    \item \textit{Score 3 (Limited Realism):} Some aspects follow physics, but there are noticeable errors in object interactions, weight distribution, or motion continuity.
    \item \textit{Score 4 (Good Physical Consistency):} Characters and objects move naturally, with only minor deviations from real-world physics.
    \item \textit{Score 5 (Flawless Adherence to Physics):} 
    The video demonstrates perfect motion realism, where all movements, interactions, and environmental effects behave.
\end{itemize}

\section{More Visualization and Analysis}

\subsection{Advantages of Hierarchical Generation}
Figure~\ref{more_visualization} presents more visualization for \Ours.
From the image, the advantages of hierarchical generation are evident in its structured storytelling and controlled content organization.
The image unfolds the narrative through Script Synopsis → Sub-Script → Scene → Shot, ensuring a clear and logical progression.
Each level incorporates emotional themes, key characters, and cinematographic details, enhancing coherence and readability. 
This structured approach not only improves narrative consistency but also strengthens visual representation. 
Particularly in AI-generated content, hierarchical generation enables precise control over story details, ensuring seamless coordination between different levels and facilitating high-quality, film-grade script generation.

\subsection{Discussion on Failure Cases and Improvements}

Figure~\ref{badcase} presents some failure cases for \Ours, reveal several key challenges in automated movie generation.
These issues span multiple aspects, including visual quality, character consistency, script faithfulness, narrative coherence, and adherence to physical laws. 
Below, we analyze the identified failures and propose potential improvements:

\textbf{Visual Appeal: Heavy Distortions and Glitches.}
The generated images exhibit severe distortions and artifacts, particularly in facial structures and animations.
These issues can break immersion and make the characters appear unnatural.
There are some potential improvements, such as further enhancing image or video generation models by using higher-quality data or incorporating a reward model to optimize outputs. 
The reward model can penalize low-quality image generations, driving the system toward producing more refined and visually appealing results.

\textbf{Character Consistency: Confusion and Weak ID Persistence.}
In several frames, character identities are inconsistent, particularly for small targets.
The same character may appear with different facial expressions, styles, or misaligned features, leading to continuity errors.
At the data level, a possible improvement strategy is to train the model with larger and higher-quality datasets that ensure better character consistency in images and videos. 
At the algorithmic level, more efficient temporal consistency mechanisms can be explored, such as tracking embeddings across frames, enforcing ID-aware latent space regularization, and implementing strict feature-matching constraints to maintain coherence across sequences.

\textbf{Script Faithfulness: Prompt Following Issues.}
Current video generation models struggle to produce complex human interactions, especially when prompts include actions like walking and talking. 
These models often fail to accurately capture and synchronize such interactions, making it challenging to generate realistic and coordinated human movements.
A potential solution is to use higher-quality data and design more efficient strategies to enhance the prompt-following ability, enabling a stronger perception~\cite{wu2023diffumask,wu2023datasetdm} of various objects and interactions in the physical world.

\textbf{Narrative Coherence: Disjointed Scene Transitions.}
The battle sequence between Shenggongbao and Nezha highlights an abrupt transition. 
Shot 1 shows Shenggongbao charging, but Shot 2 skips directly to Ne Zha preparing to defend, with no  transition that meeting to confrontation. 
Some potential improvements include refining the LLM agent by utilizing a more precise internal Chain-of-Thought, along with hierarchical storyboarding to ensure smoother transitions between actions and enhance narrative coherence.

\textbf{Physical Law Violations: Unrealistic Human Features and Proportions.}
Several frames showcase unnatural anatomical proportions and incorrect physics, such as distorted hands, limbs, and body movements that do not align with real-world constraints.
Some potential improvements include integrating physics-based rendering and leveraging biomechanical constraints to ensure that character movements and proportions align with realistic human kinematics.


{
    \small
    \bibliographystyle{ieeenat_fullname}
    \bibliography{main}
}

\end{document}